\pdfoutput=1
\documentclass[10pt,twocolumn]{article}

\usepackage[T1]{fontenc}
\usepackage[utf8]{inputenc}
\usepackage{lmodern}
\usepackage{textcomp}
\usepackage[expansion=false]{microtype}

\usepackage[letterpaper,margin=1in]{geometry}
\usepackage{xcolor}
\usepackage{graphicx}
\usepackage{amsmath,amssymb}
\usepackage{booktabs}
\usepackage{array}
\usepackage[labelsep=period]{caption}
\usepackage{xurl}
\usepackage[hidelinks]{hyperref}
\usepackage{enumitem}

\makeatletter
\renewcommand{\@seccntformat}[1]{\csname the#1\endcsname.\quad}
\renewcommand\section{\@startsection {section}{1}{\z@}%
                                   {-1.6ex \@plus -0.4ex \@minus -.2ex}%
                                   {0.5ex \@plus .2ex}%
                                   {\normalfont\large\bfseries}}
\renewcommand\subsection{\@startsection{subsection}{2}{\z@}%
                                     {-1.2ex\@plus -0.3ex \@minus -.2ex}%
                                     {0.4ex \@plus .2ex}%
                                     {\normalfont\normalsize\bfseries}}
\makeatother

\hypersetup{
  pdftitle={A Shared-Backbone Approach for Multi-Task MedMNIST Classification},
  pdfauthor={\c{S}tefan-Dorian Gavril, Andrei Arhire, Adrian Iftene}
}

\begin{document}

\twocolumn[
\begin{center}
  {\LARGE\bfseries A Shared-Backbone Approach for Multi-Task MedMNIST Classification\par}
  \vspace{0.8em}
  {\large \c{S}tefan-Dorian Gavril, Andrei Arhire, Adrian Iftene\par}
  \vspace{0.3em}
  {\normalsize Faculty of Computer Science, Alexandru Ioan Cuza University of Ia\c{s}i, Romania\par}
  \vspace{0.15em}
  {\small\ttfamily stefan.dorian.gavril@gmail.com, andrei.arhire@info.uaic.ro, adiftene@info.uaic.ro\par}
  \vspace{1.0em}
\end{center}
]

\begin{center}
  {\bfseries Abstract}
\end{center}
\begingroup
\itshape
\noindent
Multi-task biomedical classification requires models to generalize across disparate modalities and class distributions. We study 11 heterogeneous MedMNIST datasets using the harmonic mean of per-task macro-F1. We evaluate three backbones with task-specific linear heads. We identify a resolution domain shift between the MedMNIST API and evaluation environment. Resolving this inconsistency and optimizing architecture-specific regularization substantially improved performance. Our best configuration, a ConvNeXt-Tiny backbone with label smoothing, achieved a leaderboard harmonic-mean macro-F1 of 0.73294 in the Tensor Reloaded: Multi-Task MedMNIST competition, ranking sixth at the close of the official competition phase. Our implementation is publicly available online at \href{https://github.com/GavrilStefan-Dorian/A-Shared-Backbone-Approach-for-Multi-Task-MedMNIST-Classification}{\textcolor[rgb]{0,0.45,0.75}{github.com/GavrilStefan-Dorian}}.
\endgroup

\smallskip
\noindent\textbf{Keywords:} MedMNIST, multi-task learning, transfer learning, biomedical image classification, label smoothing.

\bigskip

\section{Introduction}
Multi-task learning in medical imaging allows a single model to exploit shared low-level visual features across varied modalities while maintaining task-specific classification heads. This is appealing in principle, but biomedical imaging makes it hard in practice: modalities differ wildly in appearance, dataset sizes can span several orders of magnitude, and class distributions are frequently skewed toward a dominant diagnosis. The MedMNIST collection~\cite{yang2023medmnistv2} packages this difficulty into a convenient benchmark of standardised, low-resolution biomedical datasets. It is precisely this combination of heterogeneity and small image size that makes the multi-task setting interesting to study --- enough so that it has become the basis of an active Kaggle benchmark track (Tensor Reloaded Multi-Task MedMNIST~\cite{tensorreloaded}), which we use throughout this paper. The setting is hard in two concrete ways. First, the 11 datasets used here range from 546 training images (BreastMNIST) to 165{,}466 (TissueMNIST). The harmonic mean evaluation metric means a model that ignores small tasks will score poorly regardless of how well it handles large ones. Second, the evaluation test set is fixed at 28$\times$28, while the public MedMNIST API provides higher-resolution 64px, 128px, and 224px downloads. We initially trained on native 64px API images and evaluated on 28px test images, which creates a domain shift that is easy to miss if you only look at validation scores.

This paper walks through the decisions and mistakes that led to the final result. The goal was not to find the best possible architecture, but to understand what actually moves the score on this kind of heterogeneous, harmonic-mean-scored multi-task benchmark.

\begin{figure*}[t]
\centering
\includegraphics[width=\textwidth]{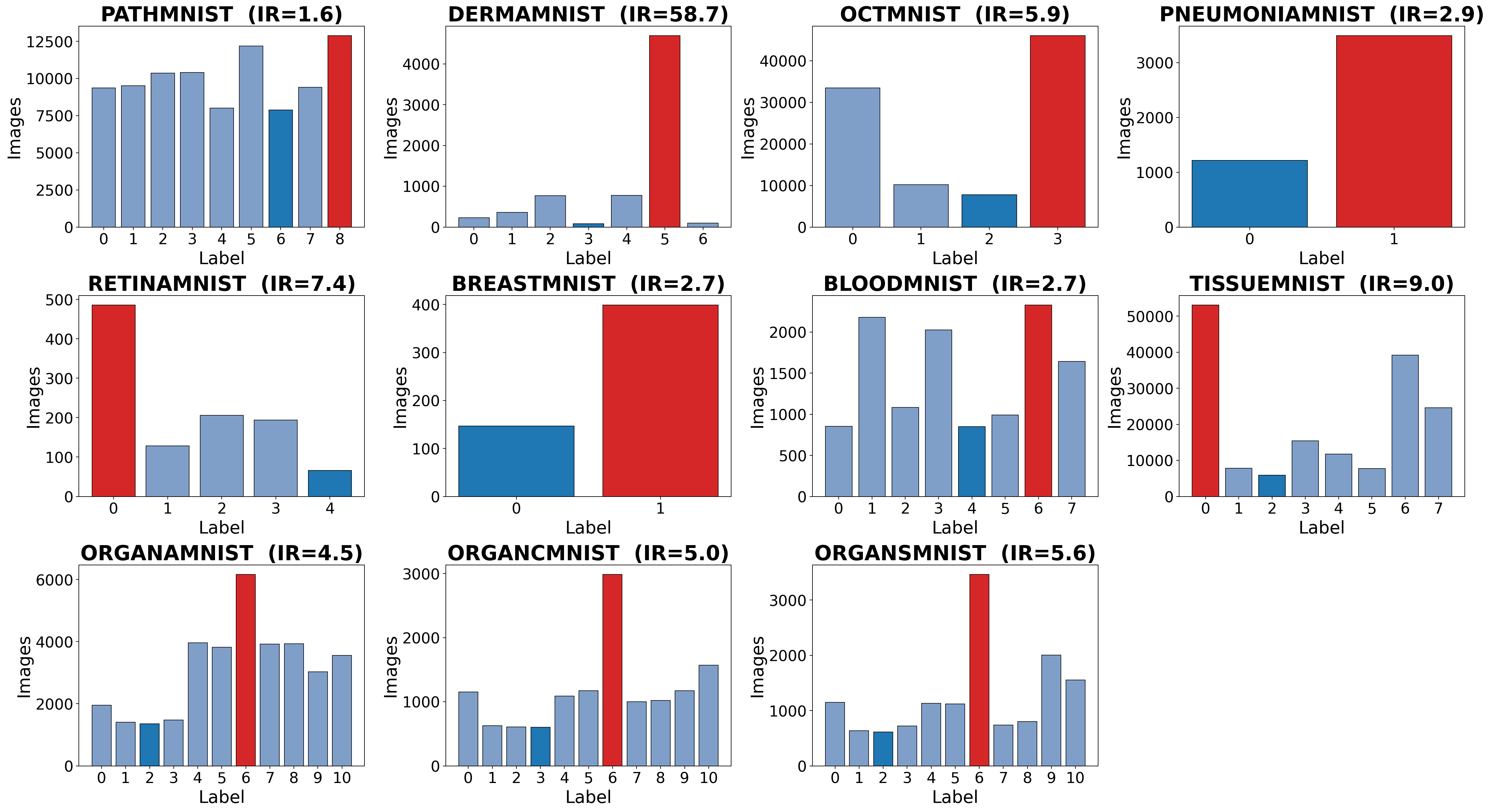}
\caption{Class distributions and imbalance ratios across the MedMNIST tasks.}
\label{fig:imbalance}
\end{figure*}
\section{Related Work}

Yang et al.~\cite{yang2021medmnist} introduced MedMNIST as a standardised lightweight benchmark for medical image classification, with all images downscaled to 28$\times$28 pixels, and benchmarked ResNet-18 and ResNet-50 at 28 and 224 pixel resolutions, reporting accuracy and AUC across 10 datasets; their results already show large variation in task difficulty, with ResNet-18 at 28px scoring 0.921 accuracy on OrganAMNIST but only 0.515 on RetinaMNIST. MedMNIST v2~\cite{yang2023medmnistv2} extended the benchmark to 12 two-dimensional datasets, while MedMNIST+~\cite{doerrich2025rethinking} later introduced 64, 128, and 224-pixel variants.

Handling this heterogeneity well depends on two ideas that are largely orthogonal to MedMNIST itself. The first is transfer learning: Raghu et al.~\cite{raghu2019transfusion} found that ImageNet pretraining offered limited performance gains on the evaluated medical imaging tasks, although it improved convergence. The second is multi-task learning: Caruana~\cite{caruana1997multitask} showed that training a shared representation across multiple related tasks acts as implicit regularisation, typically through a shared convolutional backbone with task-specific linear heads, where gradients from larger tasks help smaller tasks avoid overfitting --- directly relevant here given the 300$\times$ size difference between TissueMNIST and BreastMNIST.
Three backbones are evaluated in this study: ResNet-18~\cite{he2016resnet}, a compact residual network with 11.2M parameters that serves as the standard baseline; EfficientNet-B0~\cite{tan2019efficientnet}, a 4.1M parameter network that scales width, depth, and resolution jointly and whose MBConv blocks preserve spatial information at low input resolutions; and ConvNeXt-Tiny~\cite{liu2022convnext}, a 28M parameter modernised ConvNet that matches Vision Transformer design choices while retaining the inductive biases of convolutional architectures.

\begin{table*}[t]
\centering
\small
\caption{Dataset characteristics. Cl. = Classes; Imb. = Imbalance Ratio; Ent. = Normalized Shannon Entropy; BF1 = Baseline F1.}
\label{tab:dataset_stats}
\begin{tabular*}{\textwidth}{@{\extracolsep{\fill}}llcccccccc@{}}
\toprule
\textbf{Task} & \textbf{Modality} & \textbf{Shape} & \textbf{Cl.} & \textbf{Train} & \textbf{Val} & \textbf{Test} & \textbf{Imb.} & \textbf{Ent.} & \textbf{BF1} \\
\midrule
PathMNIST & Colon Pathology & 28x28x3 & 9 & 89,996 & 10,004 & 7,180 & 1.63 & 0.994 & 0.028 \\
DermaMNIST & Dermatoscope & 28x28x3 & 7 & 7,007 & 1,003 & 2,005 & 58.66 & 0.581 & 0.115 \\
OCTMNIST & Retinal OCT & 28x28 & 4 & 97,477 & 10,832 & 1,000 & 5.94 & 0.836 & 0.160 \\
PneumoniaMNIST & Chest X-Ray & 28x28 & 2 & 4,708 & 524 & 624 & 2.88 & 0.823 & 0.426 \\
RetinaMNIST & Fundus Camera & 28x28x3 & 5 & 1,080 & 120 & 400 & 7.36 & 0.874 & 0.124 \\
BreastMNIST & Breast Ultrasound & 28x28 & 2 & 546 & 78 & 156 & 2.71 & 0.840 & 0.422 \\
BloodMNIST & Blood Cell Micro. & 28x28x3 & 8 & 11,959 & 1,712 & 3,421 & 2.74 & 0.963 & 0.041 \\
TissueMNIST & Kidney Cortex Micro. & 28x28 & 8 & 165,466 & 23,640 & 47,280 & 9.05 & 0.868 & 0.061 \\
OrganAMNIST & Abdominal CT & 28x28 & 11 & 34,581 & 6,491 & 17,778 & 4.54 & 0.957 & 0.028 \\
OrganCMNIST & Abdominal CT & 28x28 & 11 & 13,000 & 2,392 & 8,268 & 4.98 & 0.950 & 0.034 \\
OrganSMNIST & Abdominal CT & 28x28 & 11 & 13,940 & 2,452 & 8,829 & 5.63 & 0.932 & 0.036 \\
\bottomrule
\end{tabular*}
\end{table*}

This hard-parameter-sharing pattern, where most of the network is shared and only the final layer is task-specific, is one of the two dominant multi-task strategies surveyed by Ruder~\cite{ruder2017overview}, who also discusses why a small set of large tasks can dominate gradient updates at the expense of smaller ones unless explicitly balanced; the same failure mode addressed here through task-weighted sampling. Where sampling alone is insufficient, more direct interventions on the optimisation itself have been proposed, such as projecting away conflicting per-task gradients before each update~\cite{yu2020gradientsurgery}; per-task loss weighting, discussed later as a direction for future work, follows a similar motivation applied at the loss level rather than the gradient level. More recent methods jointly balance task-loss scales and gradient magnitudes to mitigate optimization bias across tasks~\cite{lin2026dual}.
The choice of the harmonic mean over a simple average is itself a deliberate response to known pitfalls in evaluating imbalanced classification (Figure~\ref{fig:imbalance}), where aggregate accuracy or a plain average of per-class scores can hide poor performance on minority classes entirely~\cite{sokolova2009systematic}; macro-averaged F1, which weighs every class equally regardless of its frequency, is the natural per-task building block for a metric meant to penalise exactly this kind of hidden weakness. The role of input resolution has also been studied directly on this benchmark: Doerrich et al.~\cite{doerrich2025rethinking} compare CNN and Vision Transformer backbones across the MedMNIST+ collection at four input resolutions and find that supervised CNN fine-tuning at 64px or higher matches or exceeds most self-supervised and foundation-model alternatives, while gains beyond a certain resolution become inconsistent. This is consistent with the resolution-shift finding reported here: resolution is not a detail to fix once and forget, but a variable that interacts with both architecture choice and the specific train/test pipeline in ways that are easy to get wrong.

Recent studies have benchmarked foundation models on individual MedMNIST tasks~\cite{wu2025rethinking} and explored unified Transformer-based multi-task learning across all MedMNIST datasets~\cite{simionescu2025medformer}, while our work investigates an efficient shared CNN architecture optimized for balanced performance under a harmonic-mean macro-F1 objective.
\section{Dataset Overview}

\begin{figure}[htbp]
\centering
\includegraphics[width=\columnwidth]{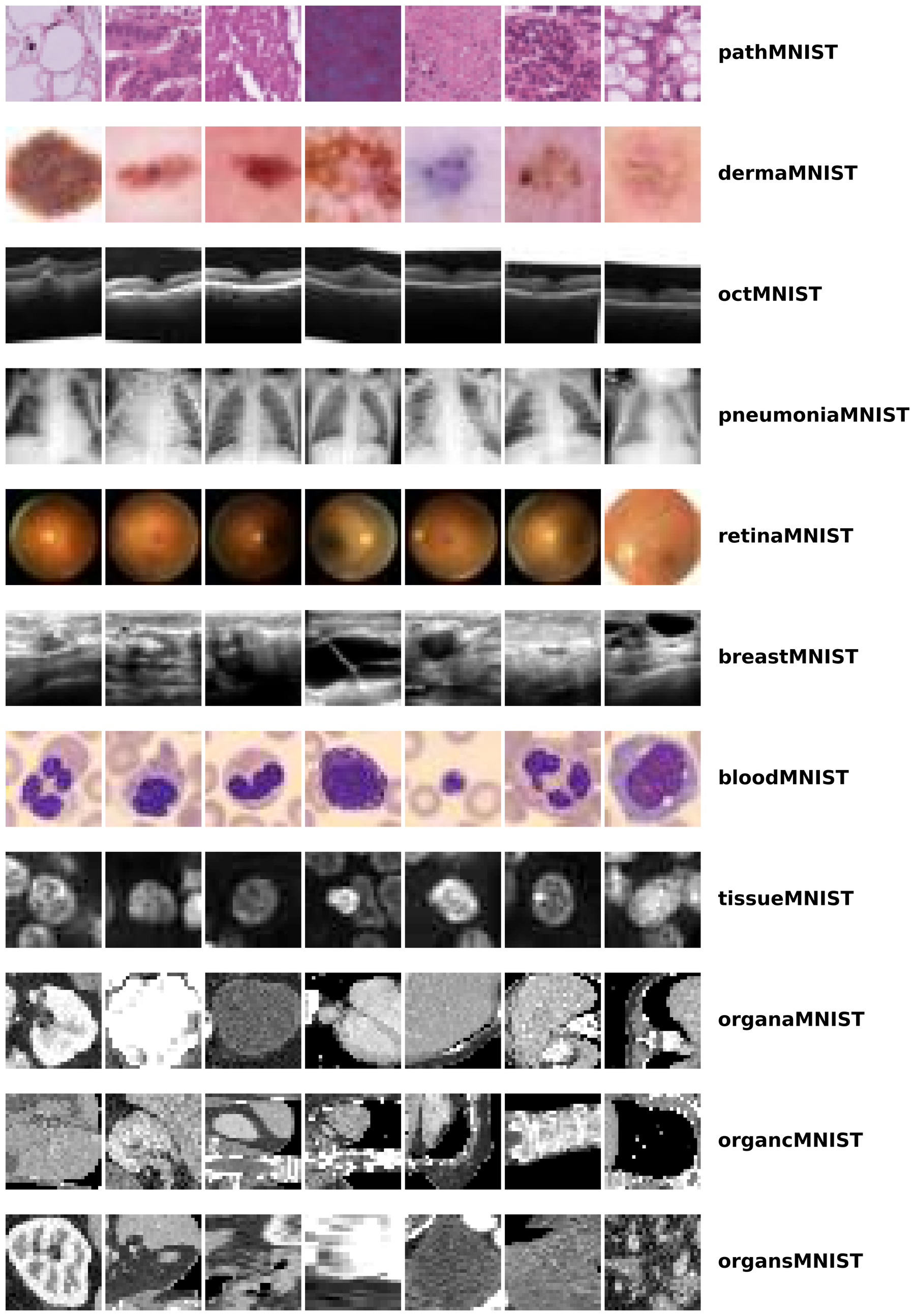}
\caption{Representative Samples from MedMNIST Datasets.}
\label{fig:representative}
\end{figure}

The MedMNIST collection contains 11 tasks with diverse modalities and class distributions. Table~\ref{tab:dataset_stats} summarizes the characteristics of these datasets, highlighting the wide range of training samples and label imbalance ratios. The datasets differ considerably in imaging modality, dataset size, number of classes, and class balance. Representative training samples from each dataset are shown in Figure~\ref{fig:representative}. Although all competition images are provided at a common resolution of $28\times28$ pixels, they retain substantial visual differences, ranging from histopathology and blood-cell microscopy to retinal fundus photography and CT imaging. This diversity motivates the use of a shared convolutional backbone capable of extracting general visual features across multiple biomedical domains. Figure~\ref{fig:imbalance} summarizes the class imbalance of each dataset. The imbalance ratio varies from nearly balanced datasets such as PathMNIST to highly skewed datasets such as DermaMNIST, whose majority class contains almost sixty times more samples than the minority class. The substantial differences in dataset size motivated task-balanced sampling to prevent larger tasks from dominating the optimization process.

\section{Method}

\subsection{Architecture}
\begin{figure*}[t]
\centering
\includegraphics[width=\textwidth]{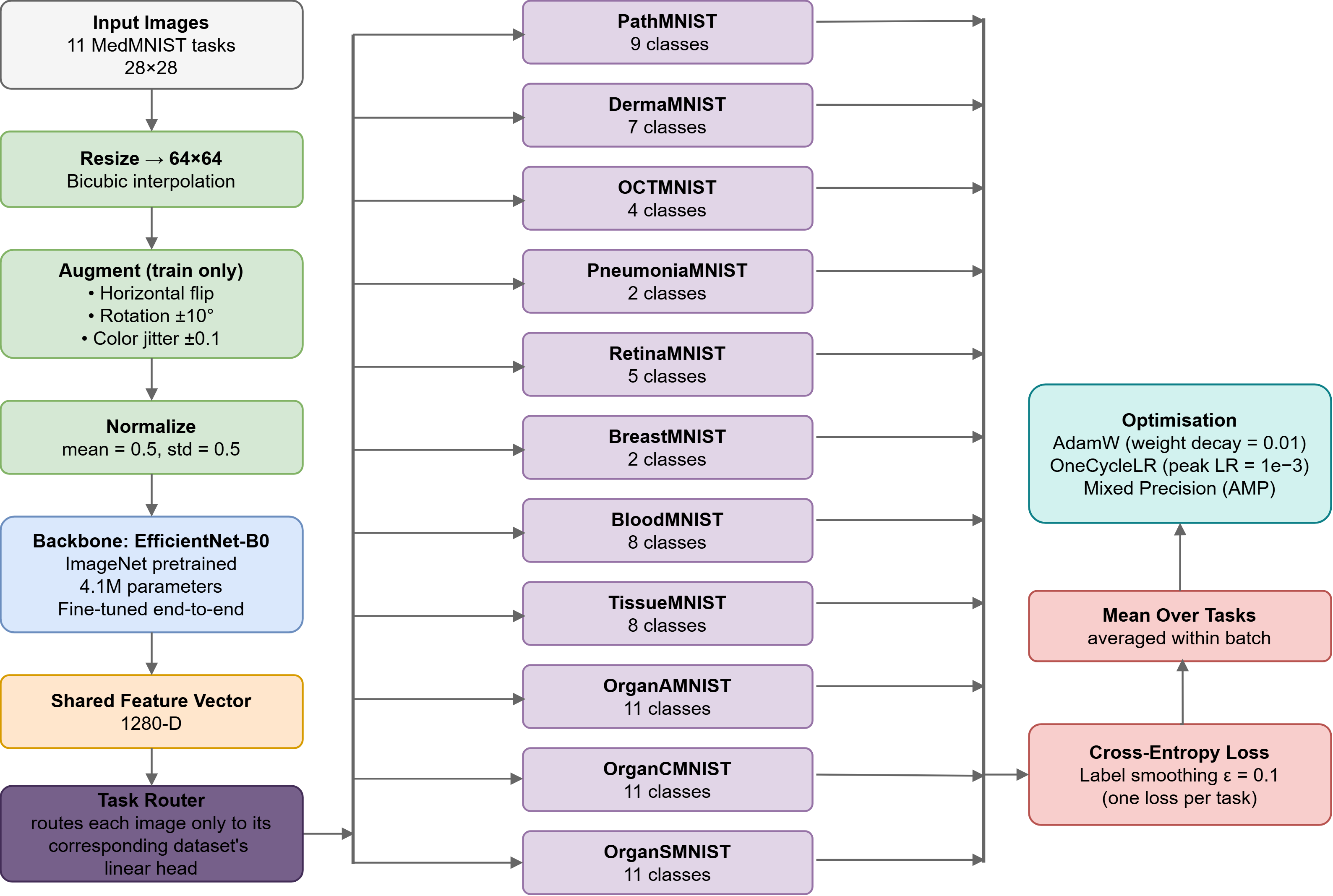}
\caption{EfficientNet-B0 Training Setup.}
\label{fig:architecture}
\end{figure*}

The model is a single shared convolutional backbone with 11 independent linear
classification heads; this work evaluates three such backbones (ResNet-18,
EfficientNet-B0, ConvNeXt-Tiny) under an identical protocol. Given an input image
$x$, the backbone extracts a feature vector $h \in \mathbb{R}^{d}$, where
$d \in \{512, 1280, 768\}$ for ResNet-18, EfficientNet-B0, and ConvNeXt-Tiny
respectively. Each task $t$ has a head
$g_t : \mathbb{R}^{d} \to \mathbb{R}^{C_t}$ where $C_t$ is the number of
classes for that task. The backbone is initialised with ImageNet pretrained weights
from the \texttt{timm} library~\cite{rw2019timm} and all parameters are fine-tuned
end-to-end. Figure~\ref{fig:architecture} illustrates the EfficientNet-B0 configuration evaluated in this study.

\subsection{Data Regime and Interpolation}
Modern backbones perform poorly at 28$\times$28 due to excessive spatial downsampling. Therefore, all Kaggle images are bicubically upscaled to 64$\times$64 within the training and test pipelines, ensuring consistent image statistics. Using native 64px MedMNIST images introduced a domain shift and reduced leaderboard performance (Section~\ref{sec:results}). Training augmentations include horizontal flipping, rotation ($\pm$10\textdegree), and colour jitter ($\pm$0.1), followed by normalization to mean 0.5 and standard deviation 0.5 per channel.

\subsection{Multi-Task Training}
At each step, a batch of images passes through the shared backbone in one forward
pass. Features are grouped by task and each group goes through its head. Per-task
cross-entropy losses are averaged:

\begin{equation}\label{eq1}
    \mathcal{L} = \frac{1}{|T_b|} \sum_{t \in T_b} \mathcal{L}_{\mathrm{CE}}^{(t)}
\end{equation}

The configuration uses AdamW (weight decay 0.01) and a OneCycleLR scheduler peaking at
$10^{-3}$. Mixed precision (AMP) is used throughout.

\subsection{Task-Balanced Sampling and Label Smoothing}
With uniform sampling, large datasets dominate every batch and small tasks get very few gradient updates. A WeightedRandomSampler assigns each sample a weight inversely proportional to the size of its corresponding dataset ($w_i=1/N_t$), ensuring that each task contributes equally in expectation. Over-sampling small datasets risks memorisation. To mitigate this, label smoothing ($\epsilon = 0.1$) is also evaluated as a standalone and combined regulariser. These two techniques are treated as independent ablation variables in Section~\ref{sec:results}, since their interaction differs per backbone.

\begin{table*}[t]
\centering
\small
\caption{Impact of WS, LS, and pretraining on the public leaderboard harmonic-mean macro-F1 score.}
\label{tab:ablation}
\begin{tabular*}{\textwidth}{@{\extracolsep{\fill}}lccccc@{}}
\toprule
\textbf{Backbone} & \textbf{Pretrained} & \textbf{WS} & \textbf{LS} & \textbf{Params} & \textbf{Public HM-F1} \\
\midrule
ConvNeXt-Tiny & Yes & False & False & 28.0M & 0.72937 \\
ConvNeXt-Tiny & Yes & False & True  & 28.0M & \textbf{0.73294} \\
ConvNeXt-Tiny & Yes & True  & False & 28.0M & 0.71892 \\
ConvNeXt-Tiny & Yes & True  & True  & 28.0M & 0.72246 \\
ConvNeXt-Tiny & No  & True  & True  & 28.0M & 0.64628 \\
\midrule
ResNet-18 & Yes & False & False & 11.2M & 0.69219 \\
ResNet-18 & Yes & False & True  & 11.2M & 0.70471 \\
ResNet-18 & Yes & True  & False & 11.2M & 0.69584 \\
ResNet-18 & Yes & True  & True  & 11.2M & 0.69464 \\
ResNet-18 & No  & True  & True  & 11.2M & 0.65394 \\
\midrule
EfficientNet-B0 & Yes & False & False & 4.1M & 0.71669 \\
EfficientNet-B0 & Yes & False & True  & 4.1M & 0.71925 \\
EfficientNet-B0 & Yes & True  & False & 4.1M & 0.72287 \\
EfficientNet-B0 & Yes & True  & True  & 4.1M & 0.70575 \\
EfficientNet-B0 & No  & True  & True  & 4.1M & 0.64196 \\
\bottomrule
\end{tabular*}
\end{table*}

\begin{table*}[t]
\centering
\footnotesize
\caption{Per-task macro F1 for the EfficientNet-B0 (pretrained, WS on, LS on)
and ConvNeXt-Tiny (pretrained, WS off, LS on) configurations.}
\label{tab:pertask}
\begin{tabular*}{\textwidth}{@{\extracolsep{\fill}}lccccccc@{}}
\toprule
\textbf{Task} & \textbf{Train} & \textbf{Val} & \textbf{Test}
& \textbf{EffNet Val} & \textbf{EffNet Test}
& \textbf{ConvNeXt Val} & \textbf{ConvNeXt Test} \\
\midrule
PathMNIST      & 89,996  & 10,004 & 7,180  & 0.978 & 0.8976 & 0.983 & 0.9061 \\
DermaMNIST     & 7,007   & 1,003  & 2,005  & 0.587 & 0.6235 & 0.636 & 0.6238 \\
OCTMNIST       & 97,477  & 10,832 & 1,000  & 0.812 & 0.6592 & 0.840 & 0.7254 \\
PneumoniaMNIST & 4,708   & 524    & 624    & 0.956 & 0.9063 & 0.966 & 0.9219 \\
RetinaMNIST    & 1,080   & 120    & 400    & 0.441 & 0.3577 & 0.433 & 0.3953 \\
BreastMNIST    & 546     & 78     & 156    & 0.834 & 0.8332 & 0.892 & 0.8301 \\
BloodMNIST     & 11,959  & 1,712  & 3,421  & 0.965 & 0.9594 & 0.970 & 0.9657 \\
TissueMNIST    & 165,466 & 23,640 & 47,280 & 0.582 & 0.5743 & 0.607 & 0.6059 \\
OrganAMNIST    & 34,581  & 6,491  & 17,778 & 0.989 & 0.9385 & 0.994 & 0.9464 \\
OrganCMNIST    & 13,000  & 2,392  & 8,268  & 0.987 & 0.9131 & 0.988 & 0.9239 \\
OrganSMNIST    & 13,940  & 2,452  & 8,829  & 0.865 & 0.7646 & 0.870 & 0.7820 \\
\bottomrule
\end{tabular*}
\end{table*}

\section{Ablation Study and Results}
\label{sec:results}

To isolate the contribution of each design choice, we compare WS and LS across three pretrained backbones and include a from-scratch WS+LS reference for each architecture; Table~\ref{tab:ablation} reports the corresponding public leaderboard harmonic-mean macro-F1 scores.

\section{Analysis}

\subsection{Pretraining establishes the baseline}
Pretrained weights substantially outperform training from scratch across all three architectures. For EfficientNet-B0 with weighted sampling and label smoothing enabled, pretraining yields a score of 0.70575 compared to 0.64196 from scratch. The pretrained models also converge much faster, reaching stable scores within the first few epochs.

\subsection{Architecture vs. Regularization}
Table~\ref{tab:ablation} shows that the optimal regularisation differs per backbone and that combining both add-ons is never the single best configuration. Both ResNet-18 (0.70471) and ConvNeXt-Tiny (0.73294) achieve their peak performance using label smoothing alone. Conversely, the EfficientNet-B0 peaks with task-balanced sampling alone (0.72287). The ConvNeXt-Tiny ultimately achieves the highest overall score, demonstrating that once the data pipeline is corrected, the increased parameter capacity of a larger backbone can be effectively leveraged.

\subsection{Fixing the domain shift}
In earlier preliminary tests (Table~\ref{tab:domain_shift}), training a ConvNeXt-Tiny model on the API's native 64px images and evaluating on the Kaggle 28px test set resulted in a substantial drop to 0.5680 due to resolution-induced distribution shift. By forcing all training data through the exact same 28px $\to$ 64px bicubic upscaling pipeline as the test data, the ConvNeXt-Tiny model recovered to a highly competitive 0.7329, indicating that data preprocessing consistency was a larger bottleneck than model capacity.
As shown in Table~\ref{tab:domain_shift}, the largest gains were observed for PathMNIST, OCTMNIST, and BreastMNIST, whereas RetinaMNIST and OrganSMNIST improved the least.

\begin{table}[htbp]
\centering
\footnotesize
\caption{Per-task test macro-F1 for ConvNeXt-Tiny using mismatched (64px train, 28px$\rightarrow$64px test) and matched (28px$\rightarrow$64px train and test) resolution pipelines.}
\label{tab:domain_shift}
\resizebox{\columnwidth}{!}{%
\begin{tabular}{lcc}
\toprule
\textbf{Task} & \textbf{Mismatched Pipeline} & \textbf{Matched Pipeline} \\
\midrule
PathMNIST      & 0.5416 & 0.9061 \\
DermaMNIST     & 0.4584 & 0.6238 \\
OCTMNIST       & 0.4675 & 0.7254 \\
PneumoniaMNIST & 0.7337 & 0.9219 \\
RetinaMNIST    & 0.3403 & 0.3953 \\
BreastMNIST    & 0.5744 & 0.8301 \\
BloodMNIST     & 0.7304 & 0.9657 \\
TissueMNIST    & 0.5055 & 0.6059 \\
OrganAMNIST    & 0.8215 & 0.9464 \\
OrganCMNIST    & 0.8482 & 0.9239 \\
OrganSMNIST    & 0.7100 & 0.7820 \\
\midrule
\textbf{Harmonic Mean} & \textbf{0.5681} & \textbf{0.7329} \\
\bottomrule
\end{tabular}%
}
\end{table}
\subsection{Per-Task Results}
Tasks split naturally into three groups based on validation F1, which was the metric used during model development and checkpoint selection. Easy tasks ($F1 > 0.90$): BloodMNIST, PathMNIST, OrganAMNIST, OrganCMNIST, PneumoniaMNIST --- large training sets with visually distinct classes. Medium ($0.70$--$0.90$): BreastMNIST, OrganSMNIST, OCTMNIST. Hard ($< 0.70$): DermaMNIST, TissueMNIST, RetinaMNIST. RetinaMNIST is consistently the worst task across every run (0.35--0.44 F1). It has only 1{,}080 training images, involves ordinal regression across 5 severity grades of diabetic retinopathy, and the diagnostic features are mostly in fine retinal structures that may be difficult to preserve at 28px. Even with balanced sampling this task is hard to improve without significantly more data or a higher native resolution. Table~\ref{tab:pertask} shows per-task macro F1 for the EfficientNet-B0 (pretrained, WS on, LS on) and the best-performing ConvNeXt-Tiny (pretrained, WS off, LS on) configurations. ConvNeXt-Tiny validation F1 values correspond to the checkpoint selected by the best validation harmonic mean. Test F1 was recomputed from this checkpoint.

\section{Discussion and Future Work}
The experiments demonstrate that data preprocessing had a larger impact than backbone capacity. Data preprocessing consistency was important in our experiments: without it, the larger ConvNeXt-Tiny model scored only 0.5680 despite its capacity advantage. Once the pipeline was corrected, backbone choice did matter as ConvNeXt-Tiny (0.73294) outperformed EfficientNet-B0 (0.72287) and ResNet-18 (0.70471). For future considerations, the most promising directions would be:
\begin{itemize}[noitemsep,topsep=2pt]
    \item \textbf{Native 64px pipeline}: training and evaluating on the API's
    native 64px images would avoid the upscaling issue entirely, at the cost
    of needing to handle the slightly different test set sizes.
    \item \textbf{Larger backbone}: a Vision Transformer or EfficientNetV2-L backbone could test whether capacity
    gains continue beyond ConvNeXt-Tiny's 28M parameters.
    \item \textbf{Per-task loss weighting}: dynamically upweighting failing tasks
    during training would better target the harmonic mean bottleneck than fixed
    sampling weights.
    \item \textbf{Test-time augmentation}: averaging predictions over multiple augmented views may improve performance without additional training.
\end{itemize}

\section{Conclusion}
This paper presents a shared-backbone multi-task classifier for the MedMNIST challenge, as well as the mistakes and fixes that led to the final result. The biggest practical finding was the train/test domain shift from mixing native API 64px images with the Kaggle 28px test set: validation scores looked fine but leaderboard performance collapsed. Fixing the data pipeline by using the same 28px$\to$64px bicubic upscaling for both training and test was an impactful change. The best submission achieved 0.73294 harmonic mean F1. RetinaMNIST remains the main bottleneck, limited by its small dataset size and ordinal label structure.

\end{document}